\pdfoutput=1

\documentclass[11pt]{article}

\usepackage[final]{acl}

\usepackage{times}
\usepackage{latexsym}
\usepackage{graphicx}
\usepackage[most]{tcolorbox}
\usepackage{cuted}

\usepackage[T1]{fontenc}

\usepackage[utf8]{inputenc}

\usepackage{microtype}

\usepackage{inconsolata}

\title{D-NLP at SemEval-2024 Task 2: Evaluating Clinical Inference Capabilities of Large Language Models}

\author{Duygu Altinok \\
  Deepgram Research, USA \\
  \texttt{duygu.altinok@deepgram.com}  \\}

\begin{document}
\maketitle
\begin{abstract}
Large language models (LLMs) have garnered significant attention and widespread usage due to their impressive performance in various tasks. However, they are not without their own set of challenges, including issues such as hallucinations, factual inconsistencies, and limitations in numerical-quantitative reasoning. Evaluating LLMs in miscellaneous reasoning tasks remains an active area of research. Prior to the breakthrough of LLMs, Transformers had already proven successful in the medical domain, effectively employed for various natural language understanding (NLU) tasks. Following this trend, LLMs have also been trained and utilized in the medical domain, raising concerns regarding factual accuracy, adherence to safety protocols, and inherent limitations. In this paper, we focus on evaluating the natural language inference capabilities of popular open-source and closed-source LLMs using clinical trial reports as the dataset. We present the performance results of each LLM and further analyze their performance on a development set, particularly focusing on challenging instances that involve medical abbreviations and require numerical-quantitative reasoning. Gemini, our leading LLM, achieved a test set F1-score of 0.748, securing the ninth position on the task scoreboard. Our work is the first of its kind, offering a thorough examination of the inference capabilities of LLMs within the medical domain.
\end{abstract}

\section{Introduction}
Large language models (LLMs) have brought about a paradigm shift in the field of Natural Language Processing (NLP) \cite{kojima2023large, wei2022emergent}. Their exceptional performance across various tasks has led to a surge in real-world applications utilizing LLM-based technology. However, a notable drawback of LLMs is their propensity to generate plausible yet incorrect information, commonly referred to as "hallucinations" \cite{ji2023}.

The remarkable breakthrough of LLMs has raised questions regarding their "intelligent" capabilities, particularly in reasoning and inference \cite{zhao2023survey, chang2023survey, laskar2023systematic}. Two specific areas that have garnered significant attention in relation to LLMs' reasoning abilities are numerical-quantitative reasoning and natural language inference. These areas are considered integral to human intelligence, prompting researchers to establish benchmarks and evaluate LLM performance in these domains \cite{stolfo2023causal, yuan2023large}. LLMs often exhibit limited performance in solving arithmetic reasoning tasks, frequently producing incorrect answers \cite{imani2023mathprompter}. Unlike natural language understanding, math problems typically possess a single correct solution, making the accurate generation of solutions more challenging for LLMs. Regarding NLI, performance reduction can be observed due to shortcut learning \cite{megnan} and hallucinations \cite{mckenna2023sources}. These investigations aim to discern whether LLMs are mere memorizers of training data or possess genuine logical reasoning abilities.

The volume of medical publications, including clinical trial data, has experienced a significant upsurge in recent years. The SemEval-2023 Task 7, known as Multi-Evidence Natural Language Inference for Clinical Trial Data (NLI4CT), aimed to address the challenge of large-scale interpretability and evidence retrieval from breast cancer clinical trial reports \cite{prev-semeval}. This task required multi-hop biomedical and numerical reasoning, which are crucial for developing systems capable of interpreting and retrieving medical evidence on a large scale, thereby facilitating personalized evidence-based care. While the previous iteration of NLI4CT resulted in the development of LLM-based models \cite{zhou-etal, kanakarajan-sankarasubbu-2023-saama, vladika-matthes-2023-sebis} achieving high performance (e.g., F1-score $\approx 85\%$), the application of LLMs in critical domains, such as real-world clinical trials, necessitates further investigation. Consequently, the second iteration of NLI4CT, SemEval-2024 Task 2, titled "Safe Biomedical Natural Language Inference for Clinical Trials" \cite{2024-semeval} is proposed, featuring an enriched dataset that includes a novel contrast set obtained through interventions applied to statements in the NLI4CT test set. Our work involves the evaluation of various popular open-source and closed-source LLMs on the development and test sets to explore their reasoning capabilities in the domain of medical NLI. We present the results by thoroughly analyzing the performance on the development set, with the best-performing LLM ranking ninth on the task leaderboard.  We have made the results on the development set available on our GitHub repository\footnote{\url{https://github.com/DuyguA/SemEval2024_NLI4CT}}.

Another aspect of our work was that we deliberately refrained from investing significant effort into prompting or experimenting with different prompts. Additionally, we aimed to showcase the remarkable development of LLMs, demonstrating their capacity to effectively engage with the task while minimizing dependence on the prompt.

\section{Related Work}
With the emergence of large language models (LLMs), there has been a growing interest in exploring their capabilities within the clinical domain. Recent studies have delved into both the potential of LLMs and the associated risks when applied in clinical settings. For instance, \cite{hung2023walking} conducted experiments utilizing GPT-3.5 on various medical NLP datasets, assessing metrics such as factuality and safety, ultimately highlighting the high level of safety offered by GPT-3.5 \footnote{\url{https://platform.openai.com/docs/models/gpt-3-5}}. \cite{pal2023medhalt} focused on the challenges posed by hallucinations in LLMs and proposed a benchmark dataset called Med-HALT (Medical Domain Hallucination Test) to evaluate popular LLMs on this front. 

Regarding the reasoning capabilities of LLMs, \cite{kwon2024large} introduced a diagnostic framework that prioritizes reasoning and employs prompt-based learning. The study specifically focused on clinical reasoning for disease diagnosis, where the LLMs generate diagnostic rationales to provide insights into patient data and the reasoning path leading to the diagnosis, known as Clinical Chain-of-Thought (Clinical CoT), using GPT-3.5 and GPT-4 \cite{openai2024gpt4}. Notably, none of the previous studies simultaneously examined the performance of both open-source and closed-source LLMs, particularly with a comprehensive focus on inference. Consequently, our work stands as the first of its kind in this regard. 

\section{Task and Dataset Description}

The clinical trials used to construct the dataset were sourced from ClinicalTrials.gov\footnote{\url{https://clinicaltrials.gov}}, a comprehensive database managed by the U.S. National Library of Medicine. ClinicalTrials.gov contains information on various clinical studies conducted worldwide, both publicly and privately funded. The dataset specifically focuses on clinical trials related to breast cancer and includes a total of 1,000 trials written in English.

\begin{itemize}
\item  \textbf{Eligibility Criteria:} This includes a set of conditions that determine the eligibility of patients to participate in the clinical trial. These criteria may include factors such as age, gender, and medical history.
\item \textbf{Intervention:} This field provides information about the type, dosage, frequency, and duration of treatments being studied within the clinical trial.
\item \textbf{Results}: The results section of each CTR reports the outcome of the trial, including data such as the number of participants, outcome measures, units of measurement, and the observed results.
\item \textbf{Adverse Events}: This field documents any unwanted side effects, signs, or symptoms observed in patients during the course of the clinical trial.
\end{itemize}

For the task at hand, each CTR may contain one or two patient groups, known as cohorts or arms, which may receive different treatments or have different baseline characteristics.

\begin{figure*}[t]
  \centering
  \includegraphics[width=\textwidth]{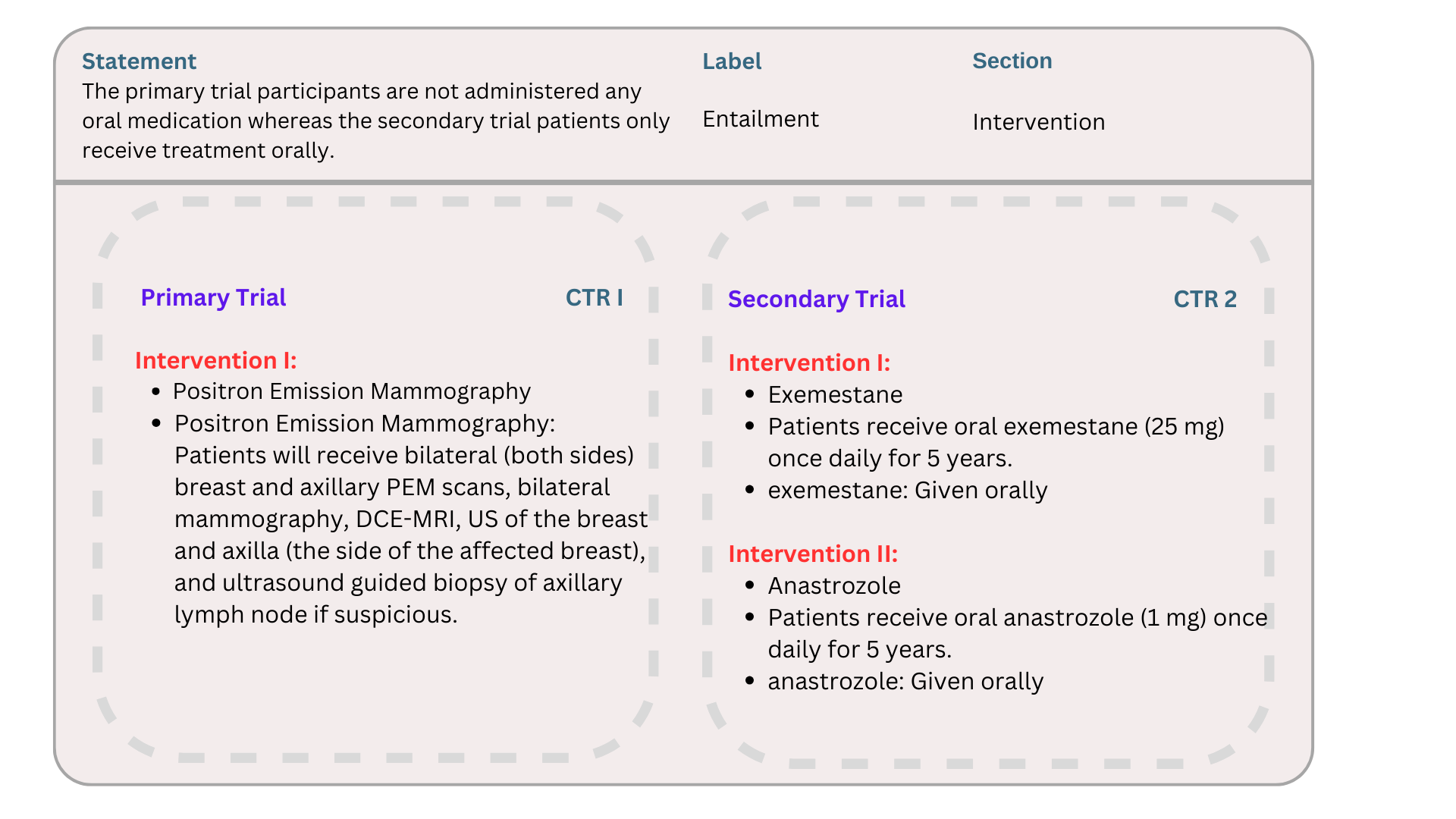}
  \caption{An example comparison task from the training set with two CTRs.}
  \label{fig:example}
\end{figure*}

The dataset consists of a total of 7,400 statements. These statements were divided into a training dataset comprising 1,700 statements, a development dataset containing 200 statements, and a hidden test dataset consisting of 5,500 statements. The statements can be categorized into two types: those that are solely related to a single CTR and others that involve a comparison between two different reports. Each statement in the dataset is labeled as either "entailment" or "contradiction". Figure \ref{fig:example} shows an example statement from the training set.

The task primarily involves binary classification, aiming to predict whether the label corresponds to entailment or contradiction. The evaluation process encompasses three aspects. Initially, the macro F1-score is computed based on the binary classification results. Subsequently, two semantic evaluations are conducted: faithfulness and consistency. Faithfulness assesses the system's ability to arrive at accurate predictions for the correct reasons, while consistency measures the system's ability to produce consistent outputs for semantically equivalent problems. The task organizers evaluate faithfulness by providing semantically altered instances, and consistency by providing preserved instances for comparison.

\section{Language Model Performance Evaluation}
This section aims to provide a detailed analysis of the performance achieved by each individual LLM. Based on the evaluation of various LLMs, including closed-source models like GPT-3.5 (ChatGPT), Claude \cite{claude-instant}, and Gemini Pro \cite{geminiteam2023gemini}, as well as open-source models like Falcon 40B \cite{falcon40b}, Mixtral 8x7B \cite{jiang2024mixtral}, and Llama 2 70B \cite{touvron2023llama}, the performance of these models was assessed on the dev and test sets. Table \ref{tab:llms} provides comprehensive information regarding the release dates and parameter sizes, measured in token size, for each LLM.

\begin{table}
\centering
\begin{tabular}{|l|l|l|l|}
\hline
\textbf{Model} & \textbf{Release Date} & \textbf{Params}\\ 
\hline
GPT-3.5 & Mar-2022 &  x \\
Claude & Mar-2023 & x \\ 
Gemini Pro & Dec-2023 & x \\ 
PaLM  & Mar-2023 & 540B \\
Falcon 40B & May-2023 & 40B \\ 
Mixtral 8x7B & Dec-2023 & 12B \\
Llama 2 70B & Jul-2023 & 130GB \\\hline
\end{tabular}
\caption{Comparison of the LLMs used in our work, indicating the parameter sizes for known closed-source LLMs and denoting unknown parameter sizes with "x".}
\label{tab:llms}
\end{table}

 All conversations took place on the Poe.com platform, providing users with a seamless chat experience. To transmit both the development set and the test set instances, we utilized an API wrapper code in a Python script, which can be accessed in our GitHub repository. As mentioned earlier, we intentionally avoided extensive prompting and instead employed a straightforward, consistent prompt for all instances. Each model's chat session commenced with a greeting, followed by a brief introductory sentence regarding the task, and subsequently, all instances were dispatched via the Python script. Appendix \ref{sec:prompts} provides information regarding the prompts.

Table \ref{tab:dev} and Table \ref{tab:test} presents a concise overview of the results obtained on the dev and test sets. Gemini Pro emerged as the best-performing model, ranking first on both the dev and test sets. Following Gemini Pro, Claude and PaLM, two closed-source LLMs, secured the second and third positions, respectively. Falcon 40B, an open-source LLM, achieved the fourth place and outperformed GPT-3.5. The last two positions were occupied by two open-source LLMs, Llama 2 70B and Mixtral 8x7B.

\begin{table}
\centering
\begin{tabular}{|l|l|l|l|l|}
\hline
\textbf{Model} & \textbf{Acc} & \textbf{F1} & \textbf{Prec} & \textbf{Recall}\\ 
\hline
Gemini Pro & 0.82 & 0.81 & 0.82 & 0.8\\
Claude & 0.81 & 0.80 & 0.81 & 0.81 \\ 
PaLM  & 0.79 & 0.78 & 0.79 & 0.79 \\
Falcon 40B & 0.745 & 0.74 & 0.74 & 0.74  \\ 
GPT-3.5 & 0.705 & 0.7 & 0.711 & 0.70  \\
Llama 2 70B & 0.675 & 0.67 & 0.68 & 0.67  \\
Mixtral 8x7B & 0.655 & 0.64 & 0.67 & 0.65 \\\hline
\end{tabular}
\caption{Accuracy, macro F1-score, precision and recall results on the development set for each LLM.}
\label{tab:dev}
\end{table}

\begin{table}
\centering
\begin{tabular}{|l|l|l|l|}
\hline
\textbf{Model} & \textbf{F1} & \textbf{Faith} & \textbf{Consist}\\ 
\hline
Gemini Pro & 0.75 & 0.83 & 0.74 \\
Claude & 0.73 & 0.83 & 0.72 \\ 
PaLM  & 0.72 & 0.87 & 0.73 \\
Falcon 40B & 0.702 & 0.569 & 0.609 \\ 
GPT-3.5 & 0.684 & 0.74 & 0.66 \\
Llama 2 70B & 0.682 & 0.693 & 0.638 \\
Mixtral 8x7B & 0.604 & 0.899 & 0.73 \\\hline
\end{tabular}
\caption{Macro F1-score, faithfulness and consistency results on the test set for each LLM.}
\label{tab:test}
\end{table}

In the next section, we delve into the detailed performance analysis of the language models on the development set, focusing on specific cases of interest.

\subsection{General Performance Evaluation}
Among the top-ranking LLMs, namely Gemini Pro, Claude, PaLM, and Falcon 40B, their performance on the development set was indeed remarkable. The number of inaccurate predictions made by each LLM on the development set of 200 instances is presented in Table \ref{tab:incorrects}. There were only 3 instances in the development set that were incorrectly predicted by all LLMs.

\begin{table}
\centering
\begin{tabular}{|l|l|}
\hline
\textbf{Model} & \textbf{Incorrect} \\ 
\hline
Gemini Pro & 36\\
Claude & 38\\ 
PaLM  & 42\\
Falcon 40B & 51\\ 
GPT-3.5 & 59 \\
Llama 2 70B & 65 \\
Mixtral 8x7B & 69 \\\hline
\end{tabular}
\caption{Number of incorrect predictions on the development set of 200 instances for each LLM.}
\label{tab:incorrects}
\end{table}

Among the top-performing LLMs, a set of 12 instances emerged as particularly challenging, denoted as "difficult instances". These instances present a significant challenge, as none of the top three performer LLMs in the set - Gemini Pro, Claude and PaLM, were able to accurately predict their outcomes. Within this subset, 2 instances involved medical abbreviations, 3 featured numerical entities, and the remaining 7 were categorized as "plain instances". The subsequent subsections will delve into a detailed analysis of the difficult instances encompassing numerical entities and abbreviations. For the present discussion, our attention will be directed towards the plain difficult instances.

\begin{figure*}[t]
  \centering
  \includegraphics[width=\textwidth]{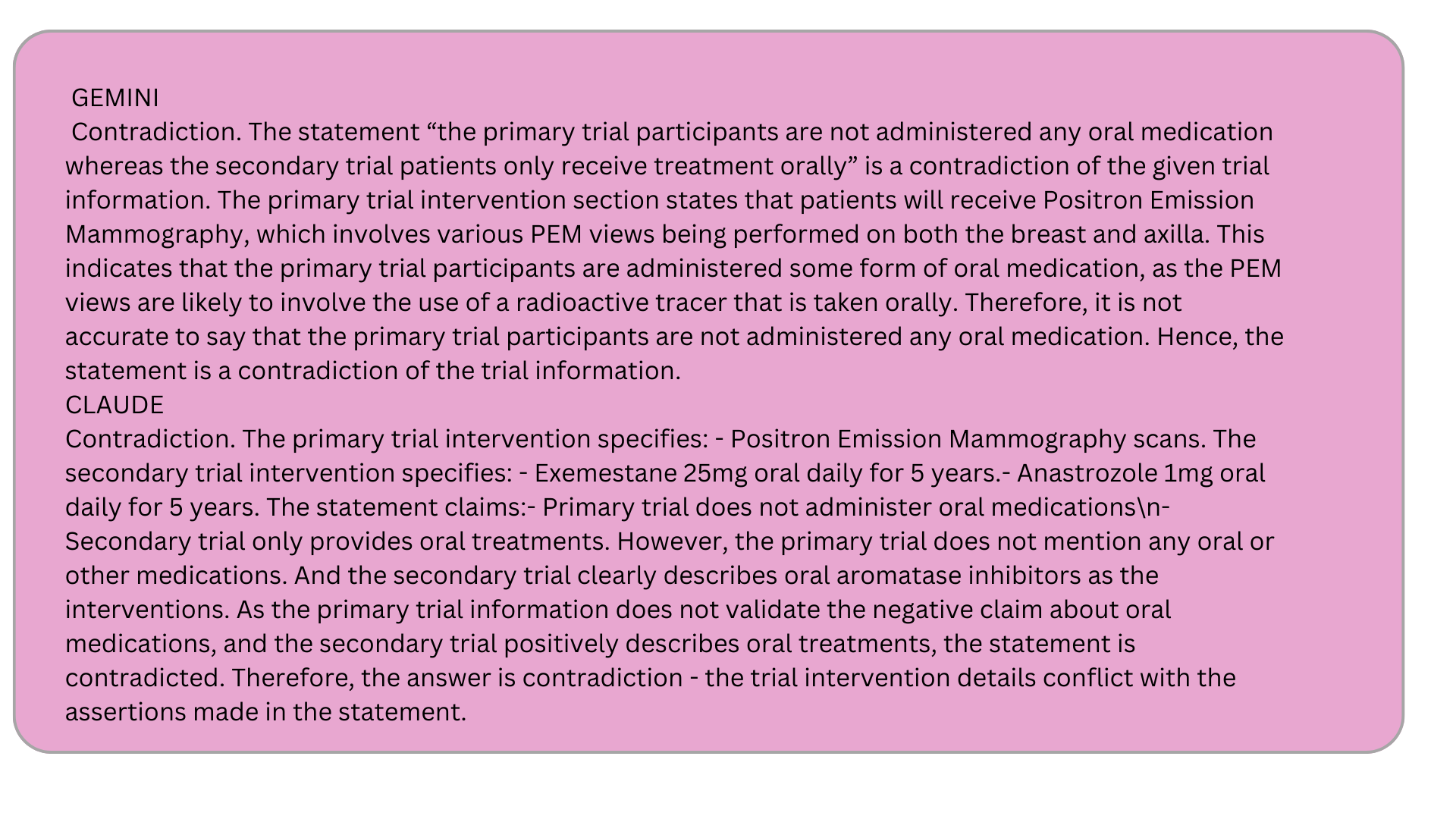}
  \caption{An example comparison task from the training set with two CTRs.}
  \label{fig:gemini-claude-general}
\end{figure*}

One particular difficult instance is illustrated in Figure \ref{fig:example}, where the top two performing LLMs provided the answers given by Figure \ref{fig:gemini-claude-general}.

Gemini made a somewhat fair mistake by incorporating knowledge from the training data, suggesting that PET includes orally administered tracer drugs. However, it should be noted that tracer drugs do not qualify as any form of medication for treating cancer or other diseases. In this case, Gemini falters by "assuming too much" and "relying too heavily on prior knowledge." It is important to clarify that this is not a hallucination. On the other hand, Claude correctly deduced that there were no oral medications in the primary trial and that oral medication was administered in the secondary trial. However, Claude failed to synthesize this information and draw a conclusion, indicating a breakdown in reasoning from evidence. Similarly, PaLM and Falcon 40B encountered the same issue as Claude. While they accurately pointed out the evidence, they failed in the reasoning process. In the case of PaLM, it did not provide any indications of engaging in reasoning. Falcon 40B made a literal mistake by becoming mired in the intricacies of the language. Its answer includes the statement, "the statement cannot be directly entailed from the intervention information provided. Specifically, while the primary trial does not mention oral medication, the secondary trial does not exclusively mention oral medication, contradicting the statement."

These mistakes range from overthinking, failure to follow the evidence, inadequate reasoning, to becoming excessively focused on minute details—a clear manifestation of the inherent challenges that LLMs face. The less performing LLMs, namely GPT-3.5, LLama 2 70B, and Mixtral 7x8B, demonstrated a decent performance considering the task difficulty. However, they exhibited a relatively higher frequency of failures in reasoning and inference compared to the top-performing LLMs.

Having examined the plain difficult instances, we now turn our attention to evaluating the performance of the LLMs on instances containing medical abbreviations.

\subsection{Abbreviated Instances Performance Evaluation}
In our development set, we identified 31 instances that contained medical abbreviations. We used the ScispaCy package's abbreviation detector to extract these instances.

Among the top performers, Gemini, Claude, PaLM, and Falcon 40B made 4, 6, 7, and 8 mistakes, respectively, in handling these abbreviations. The bottom performers, GPT-3.5, Llama 2 70B, and Mixtral 8x70B, made 10, 8, and 8 mistakes, respectively.

Upon closer examination, we found that all of the LLMs were able to correctly resolve the meanings of the medical abbreviations. However, they made mistakes due to other reasoning problems.  

For example, none of Gemini's four mistakes in handling abbreviations were related to resolving their meanings. Similarly, the other LLMs also failed the task primarily due to quantitative-numerical reasoning failures. Appendix \ref{sec:examples} showcases a more comprehensive example of this particular type of occurrence and the corresponding failure.

Overall, the performance of all LLMs in resolving abbreviations was commendable. However, as mentioned before, the majority of failures stemmed from challenges in numerical-quantitative reasoning.

\subsection{Numerical Instances Performance Evaluation}
Our development set contained 78 instances with numerical entities. We employed the spaCy package \cite{spacy2} and its NER component to identify these entities. To ensure comprehensive semantic evaluation, we combined ScispaCy and spaCy models.

Among the top-performing LLMs, Gemini, Claude, PaLM, and Falcon 40B made 13, 14, 18, and 19 mistakes, respectively. Notably, the bottom performers, GPT-3.5, Llama 2 70B, and Mixtral 8x70B, made significantly more mistakes (21, 26, and 24, respectively).

Interestingly, the top performers, Gemini and Claude, made the same mistakes on numerical instances in the development set. Upon examining their responses, we observed that they performed arithmetic operations and reasoned based on the calculated results. In Appendix B, Figure \ref{fig:numeric-top-answers} and Figure \ref{fig:numeric-bottom-answers} portray instances of successful outcomes achieved by our LLMs. These figures demonstrate accurate performance in arithmetic operations and logical deduction.

However, even the top performers made occasional errors. For instance, Gemini provided an incorrect answer where there was no evidence of arithmetic operations or reasoning: "The primary trial adverse events section shows that there were 10 patients in cohort1 who suffered adverse events out of a total of 67 patients. Therefore, it is accurate to say that over 1/6 patients in cohort1 of the primary trial suffered adverse events." This suggests that Gemini did not calculate 1/6 of the total number of patients (67).

Among the numerical instances incorrectly predicted by Gemini and Claude, we found no instances where arithmetic calculations were performed. Conversely, correctly predicted instances, such as the one shown in Figure \ref{fig:semantic-parse} in Appendix \ref{sec:examples}, involved at least one mathematical operation that was logically connected to the rest of the argument. We determined these logical connections by analyzing the dependency tree of the answers, as explained in Appendix \ref{sec:examples}. Our findings indicate that when LLMs demonstrate signs of performing arithmetic operations, their results are generally reliable. Conversely, when there is no evidence of arithmetic operations, the result is likely incorrect.

The other top performers, PaLM and Falcon 40B, exhibited similar behavior to Gemini and Claude. They performed arithmetic operations and made deductions based on those operations. When they failed, they did not provide any numerical clues.

The bottom performer, GPT-3.5 was able to perform arithmetic operations. However, it struggled with simple quantity comparisons, such as n<m for random integers. Mixtral 8x7B also faced similar challenges.

Llama 2 70B performed particularly poorly on numerical instances. For the example in Figure \ref{fig:numeric-bottom-answers}, where other LLMs succeeded by performing arithmetic operations, Llama failed completely. It provided an incorrect answer without any evidence of subtraction or comparison. In fact, Llama generally struggled with numerical examples, succeeding primarily in quantitative comparisons where operands were provided directly in the context without requiring mathematical processing.

In conclusion, while other LLMs demonstrated proficiency in handling numerical entities, Llama 2 70B failed to meet expectations.

\section{Conclusion}
Our detailed analysis of LLMs' performance on various reasoning tasks in the medical domain reveals that they are not merely passive memorizers. They possess the ability to perform numerical-quantitative reasoning, general reasoning, and abbreviation resolution, even in a highly specialized domain with unique vocabulary. Notably, Falcon 40B, an open-source LLM, demonstrated impressive performance, rivaling top closed-source LLMs.

Despite their successes, LLMs are not without limitations. Occasional nonsensical predictions highlight the need for caution when using them in high-stakes domains such as medicine. However, the results of our study are highly promising and suggest that with increased training data and computational power, LLMs have the potential to become invaluable tools in the medical field.

The future of LLMs in medicine holds exciting possibilities. As these models continue to evolve, we anticipate that they will play an increasingly significant role in healthcare, transforming the way we diagnose, treat, and prevent diseases.

\section{Limitations}
As mentioned in earlier sections, we utilized the Poe platform for interacting with LLMs. All the work was accomplished within the confines of a monthly subscription fee of \$20. The results of GPT-4 are not included in this study due to the messaging limit imposed by the platform, which was exceeded by the number of instances in the test set.

\bibliography{acl_latex}

\begin{thebibliography}{26}
\expandafter\ifx\csname natexlab\endcsname\relax\def\natexlab#1{#1}\fi

\bibitem[{Almazrouei et~al.(2023)Almazrouei, Alobeidli, Alshamsi, Cappelli, Cojocaru, Debbah, Goffinet, Heslow, Launay, Malartic, Noune, Pannier, and Penedo}]{falcon40b}
Ebtesam Almazrouei, Hamza Alobeidli, Abdulaziz Alshamsi, Alessandro Cappelli, Ruxandra Cojocaru, Merouane Debbah, Etienne Goffinet, Daniel Heslow, Julien Launay, Quentin Malartic, Badreddine Noune, Baptiste Pannier, and Guilherme Penedo. 2023.
\newblock {Falcon-40B}: an open large language model with state-of-the-art performance.

\bibitem[{Anthropic(2023)}]{claude-instant}
Anthropic. 2023.
\newblock \href {https://www.anthropic.com/news/introducing-claude} {Introducing claude}.

\bibitem[{Chang et~al.(2023)Chang, Wang, Wang, Wu, Yang, Zhu, Chen, Yi, Wang, Wang, Ye, Zhang, Chang, Yu, Yang, and Xie}]{chang2023survey}
Yupeng Chang, Xu~Wang, Jindong Wang, Yuan Wu, Linyi Yang, Kaijie Zhu, Hao Chen, Xiaoyuan Yi, Cunxiang Wang, Yidong Wang, Wei Ye, Yue Zhang, Yi~Chang, Philip~S. Yu, Qiang Yang, and Xing Xie. 2023.
\newblock \href {http://arxiv.org/abs/2307.03109} {A survey on evaluation of large language models}.

\bibitem[{Du et~al.(2023)Du, He, Zou, Tao, and Hu}]{megnan}
Mengnan Du, Fengxiang He, Na~Zou, Dacheng Tao, and Xia Hu. 2023.
\newblock \href {https://doi.org/10.1145/3596490} {Shortcut learning of large language models in natural language understanding}.
\newblock \emph{Commun. ACM}, 67(1):110–120.

\bibitem[{{Gemini Team}(2023)}]{geminiteam2023gemini}
{Gemini Team}. 2023.
\newblock \href {http://arxiv.org/abs/2312.11805} {Gemini: A family of highly capable multimodal models}.

\bibitem[{Honnibal and Montani(2017)}]{spacy2}
Matthew Honnibal and Ines Montani. 2017.
\newblock {spaCy 2}: Natural language understanding with {B}loom embeddings, convolutional neural networks and incremental parsing.
\newblock To appear.

\bibitem[{Hung et~al.(2023)Hung, Rim, Frost, Bruckner, and Lawrence}]{hung2023walking}
Chia-Chien Hung, Wiem~Ben Rim, Lindsay Frost, Lars Bruckner, and Carolin Lawrence. 2023.
\newblock \href {http://arxiv.org/abs/2311.14966} {Walking a tightrope -- evaluating large language models in high-risk domains}.

\bibitem[{Imani et~al.(2023)Imani, Du, and Shrivastava}]{imani2023mathprompter}
Shima Imani, Liang Du, and Harsh Shrivastava. 2023.
\newblock \href {http://arxiv.org/abs/2303.05398} {Mathprompter: Mathematical reasoning using large language models}.

\bibitem[{Ji et~al.(2023)Ji, Lee, Frieske, Yu, Su, Xu, Ishii, Bang, Madotto, and Fung}]{ji2023}
Ziwei Ji, Nayeon Lee, Rita Frieske, Tiezheng Yu, Dan Su, Yan Xu, Etsuko Ishii, Ye~Jin Bang, Andrea Madotto, and Pascale Fung. 2023.
\newblock \href {https://doi.org/10.1145/3571730} {Survey of hallucination in natural language generation}.
\newblock \emph{ACM Comput. Surv.}, 55(12).

\bibitem[{Jiang et~al.(2024)Jiang, Sablayrolles, Roux, Mensch, Savary, Bamford, Chaplot, de~las Casas, Hanna, Bressand, Lengyel, Bour, Lample, Lavaud, Saulnier, Lachaux, Stock, Subramanian, Yang, Antoniak, Scao, Gervet, Lavril, Wang, Lacroix, and Sayed}]{jiang2024mixtral}
Albert~Q. Jiang, Alexandre Sablayrolles, Antoine Roux, Arthur Mensch, Blanche Savary, Chris Bamford, Devendra~Singh Chaplot, Diego de~las Casas, Emma~Bou Hanna, Florian Bressand, Gianna Lengyel, Guillaume Bour, Guillaume Lample, Lélio~Renard Lavaud, Lucile Saulnier, Marie-Anne Lachaux, Pierre Stock, Sandeep Subramanian, Sophia Yang, Szymon Antoniak, Teven~Le Scao, Théophile Gervet, Thibaut Lavril, Thomas Wang, Timothée Lacroix, and William~El Sayed. 2024.
\newblock \href {http://arxiv.org/abs/2401.04088} {Mixtral of experts}.

\bibitem[{Jullien et~al.(2024)Jullien, Valentino, and Freitas}]{2024-semeval}
Ma{\"e}l Jullien, Marco Valentino, and Andr{\'e} Freitas. 2024.
\newblock {S}em{E}val-2024 task 2: Safe biomedical natural language inference for clinical trials.
\newblock In \emph{Proceedings of the 18th International Workshop on Semantic Evaluation (SemEval-2024)}. Association for Computational Linguistics.

\bibitem[{Jullien et~al.(2023)Jullien, Valentino, Frost, O{'}regan, Landers, and Freitas}]{prev-semeval}
Ma{\"e}l Jullien, Marco Valentino, Hannah Frost, Paul O{'}regan, Donal Landers, and Andr{\'e} Freitas. 2023.
\newblock \href {https://doi.org/10.18653/v1/2023.semeval-1.307} {{S}em{E}val-2023 task 7: Multi-evidence natural language inference for clinical trial data}.
\newblock In \emph{Proceedings of the 17th International Workshop on Semantic Evaluation (SemEval-2023)}, pages 2216--2226, Toronto, Canada. Association for Computational Linguistics.

\bibitem[{Kanakarajan and Sankarasubbu(2023)}]{kanakarajan-sankarasubbu-2023-saama}
Kamal~Raj Kanakarajan and Malaikannan Sankarasubbu. 2023.
\newblock \href {https://doi.org/10.18653/v1/2023.semeval-1.137} {Saama {AI} research at {S}em{E}val-2023 task 7: Exploring the capabilities of flan-t5 for multi-evidence natural language inference in clinical trial data}.
\newblock In \emph{Proceedings of the 17th International Workshop on Semantic Evaluation (SemEval-2023)}, pages 995--1003, Toronto, Canada. Association for Computational Linguistics.

\bibitem[{Kojima et~al.(2023)Kojima, Gu, Reid, Matsuo, and Iwasawa}]{kojima2023large}
Takeshi Kojima, Shixiang~Shane Gu, Machel Reid, Yutaka Matsuo, and Yusuke Iwasawa. 2023.
\newblock \href {http://arxiv.org/abs/2205.11916} {Large language models are zero-shot reasoners}.

\bibitem[{Kwon et~al.(2024)Kwon, iunn Ong, Kang, Moon, Lee, Hwang, Sim, Sohn, Lee, and Yeo}]{kwon2024large}
Taeyoon Kwon, Kai~Tzu iunn Ong, Dongjin Kang, Seungjun Moon, Jeong~Ryong Lee, Dosik Hwang, Yongsik Sim, Beomseok Sohn, Dongha Lee, and Jinyoung Yeo. 2024.
\newblock \href {http://arxiv.org/abs/2312.07399} {Large language models are clinical reasoners: Reasoning-aware diagnosis framework with prompt-generated rationales}.

\bibitem[{Laskar et~al.(2023)Laskar, Bari, Rahman, Bhuiyan, Joty, and Huang}]{laskar2023systematic}
Md~Tahmid~Rahman Laskar, M~Saiful Bari, Mizanur Rahman, Md~Amran~Hossen Bhuiyan, Shafiq Joty, and Jimmy~Xiangji Huang. 2023.
\newblock \href {http://arxiv.org/abs/2305.18486} {A systematic study and comprehensive evaluation of chatgpt on benchmark datasets}.

\bibitem[{McKenna et~al.(2023)McKenna, Li, Cheng, Hosseini, Johnson, and Steedman}]{mckenna2023sources}
Nick McKenna, Tianyi Li, Liang Cheng, Mohammad~Javad Hosseini, Mark Johnson, and Mark Steedman. 2023.
\newblock \href {http://arxiv.org/abs/2305.14552} {Sources of hallucination by large language models on inference tasks}.

\bibitem[{OpenAI(2024)}]{openai2024gpt4}
OpenAI. 2024.
\newblock \href {http://arxiv.org/abs/2303.08774} {Gpt-4 technical report}.

\bibitem[{Pal et~al.(2023)Pal, Umapathi, and Sankarasubbu}]{pal2023medhalt}
Ankit Pal, Logesh~Kumar Umapathi, and Malaikannan Sankarasubbu. 2023.
\newblock \href {http://arxiv.org/abs/2307.15343} {Med-halt: Medical domain hallucination test for large language models}.

\bibitem[{Stolfo et~al.(2023)Stolfo, Jin, Shridhar, Schölkopf, and Sachan}]{stolfo2023causal}
Alessandro Stolfo, Zhijing Jin, Kumar Shridhar, Bernhard Schölkopf, and Mrinmaya Sachan. 2023.
\newblock \href {http://arxiv.org/abs/2210.12023} {A causal framework to quantify the robustness of mathematical reasoning with language models}.

\bibitem[{Touvron et~al.(2023)Touvron, Martin, Stone, Albert, Almahairi, Babaei, Bashlykov, Batra, Bhargava, Bhosale, Bikel, Blecher, Ferrer, Chen, Cucurull, Esiobu, Fernandes, Fu, Fu, Fuller, Gao, Goswami, Goyal, Hartshorn, Hosseini, Hou, Inan, Kardas, Kerkez, Khabsa, Kloumann, Korenev, Koura, Lachaux, Lavril, Lee, Liskovich, Lu, Mao, Martinet, Mihaylov, Mishra, Molybog, Nie, Poulton, Reizenstein, Rungta, Saladi, Schelten, Silva, Smith, Subramanian, Tan, Tang, Taylor, Williams, Kuan, Xu, Yan, Zarov, Zhang, Fan, Kambadur, Narang, Rodriguez, Stojnic, Edunov, and Scialom}]{touvron2023llama}
Hugo Touvron, Louis Martin, Kevin Stone, Peter Albert, Amjad Almahairi, Yasmine Babaei, Nikolay Bashlykov, Soumya Batra, Prajjwal Bhargava, Shruti Bhosale, Dan Bikel, Lukas Blecher, Cristian~Canton Ferrer, Moya Chen, Guillem Cucurull, David Esiobu, Jude Fernandes, Jeremy Fu, Wenyin Fu, Brian Fuller, Cynthia Gao, Vedanuj Goswami, Naman Goyal, Anthony Hartshorn, Saghar Hosseini, Rui Hou, Hakan Inan, Marcin Kardas, Viktor Kerkez, Madian Khabsa, Isabel Kloumann, Artem Korenev, Punit~Singh Koura, Marie-Anne Lachaux, Thibaut Lavril, Jenya Lee, Diana Liskovich, Yinghai Lu, Yuning Mao, Xavier Martinet, Todor Mihaylov, Pushkar Mishra, Igor Molybog, Yixin Nie, Andrew Poulton, Jeremy Reizenstein, Rashi Rungta, Kalyan Saladi, Alan Schelten, Ruan Silva, Eric~Michael Smith, Ranjan Subramanian, Xiaoqing~Ellen Tan, Binh Tang, Ross Taylor, Adina Williams, Jian~Xiang Kuan, Puxin Xu, Zheng Yan, Iliyan Zarov, Yuchen Zhang, Angela Fan, Melanie Kambadur, Sharan Narang, Aurelien Rodriguez, Robert Stojnic, Sergey Edunov, and Thomas
  Scialom. 2023.
\newblock \href {http://arxiv.org/abs/2307.09288} {Llama 2: Open foundation and fine-tuned chat models}.

\bibitem[{Vladika and Matthes(2023)}]{vladika-matthes-2023-sebis}
Juraj Vladika and Florian Matthes. 2023.
\newblock \href {https://doi.org/10.18653/v1/2023.semeval-1.257} {Sebis at {S}em{E}val-2023 task 7: A joint system for natural language inference and evidence retrieval from clinical trial reports}.
\newblock In \emph{Proceedings of the 17th International Workshop on Semantic Evaluation (SemEval-2023)}, pages 1863--1870, Toronto, Canada. Association for Computational Linguistics.

\bibitem[{Wei et~al.(2022)Wei, Tay, Bommasani, Raffel, Zoph, Borgeaud, Yogatama, Bosma, Zhou, Metzler, Chi, Hashimoto, Vinyals, Liang, Dean, and Fedus}]{wei2022emergent}
Jason Wei, Yi~Tay, Rishi Bommasani, Colin Raffel, Barret Zoph, Sebastian Borgeaud, Dani Yogatama, Maarten Bosma, Denny Zhou, Donald Metzler, Ed~H. Chi, Tatsunori Hashimoto, Oriol Vinyals, Percy Liang, Jeff Dean, and William Fedus. 2022.
\newblock \href {http://arxiv.org/abs/2206.07682} {Emergent abilities of large language models}.

\bibitem[{Yuan et~al.(2023)Yuan, Yuan, Tan, Wang, and Huang}]{yuan2023large}
Zheng Yuan, Hongyi Yuan, Chuanqi Tan, Wei Wang, and Songfang Huang. 2023.
\newblock \href {http://arxiv.org/abs/2304.02015} {How well do large language models perform in arithmetic tasks?}

\bibitem[{Zhao et~al.(2023)Zhao, Zhou, Li, Tang, Wang, Hou, Min, Zhang, Zhang, Dong, Du, Yang, Chen, Chen, Jiang, Ren, Li, Tang, Liu, Liu, Nie, and Wen}]{zhao2023survey}
Wayne~Xin Zhao, Kun Zhou, Junyi Li, Tianyi Tang, Xiaolei Wang, Yupeng Hou, Yingqian Min, Beichen Zhang, Junjie Zhang, Zican Dong, Yifan Du, Chen Yang, Yushuo Chen, Zhipeng Chen, Jinhao Jiang, Ruiyang Ren, Yifan Li, Xinyu Tang, Zikang Liu, Peiyu Liu, Jian-Yun Nie, and Ji-Rong Wen. 2023.
\newblock \href {http://arxiv.org/abs/2303.18223} {A survey of large language models}.

\bibitem[{Zhou et~al.(2023)Zhou, Jin, Li, Li, Liu, You, and Wu}]{zhou-etal}
Yuxuan Zhou, Ziyu Jin, Meiwei Li, Miao Li, Xien Liu, Xinxin You, and Ji~Wu. 2023.
\newblock \href {https://doi.org/10.18653/v1/2023.semeval-1.234} {{TH}i{FLY} research at {S}em{E}val-2023 task 7: A multi-granularity system for {CTR}-based textual entailment and evidence retrieval}.
\newblock In \emph{Proceedings of the 17th International Workshop on Semantic Evaluation (SemEval-2023)}, pages 1681--1690, Toronto, Canada. Association for Computational Linguistics.

\end{thebibliography}

\appendix

\section{Prompts}
\label{sec:prompts}
 We employed two prompts, the prompts for the individual task and comparison tasks outlined as follows:
 
 "Below find section\_name section of the primary trial of a clinical trial. Infer if the following statement entails from the given trial information. Answer should be either entailment or contradiction. Please justify the answer based on numbers. PRIMARY TRIAL section\_name: trial\_value STATEMENT: statement"

 "Below find section\_name sections of a primary trial and a secondary trial belonging to same clinical trial. Infer if the following statement entails from the given trial information. Answer should be either entailment or contradiction. Please justify the answer based on numbers. PRIMARY TRIAL section\_name: trial\_value1 SECONDARY TRIAL section\_name: trial\_value2 STATEMENT: statement".

\begin{figure}[t]
  \includegraphics[width=\columnwidth]{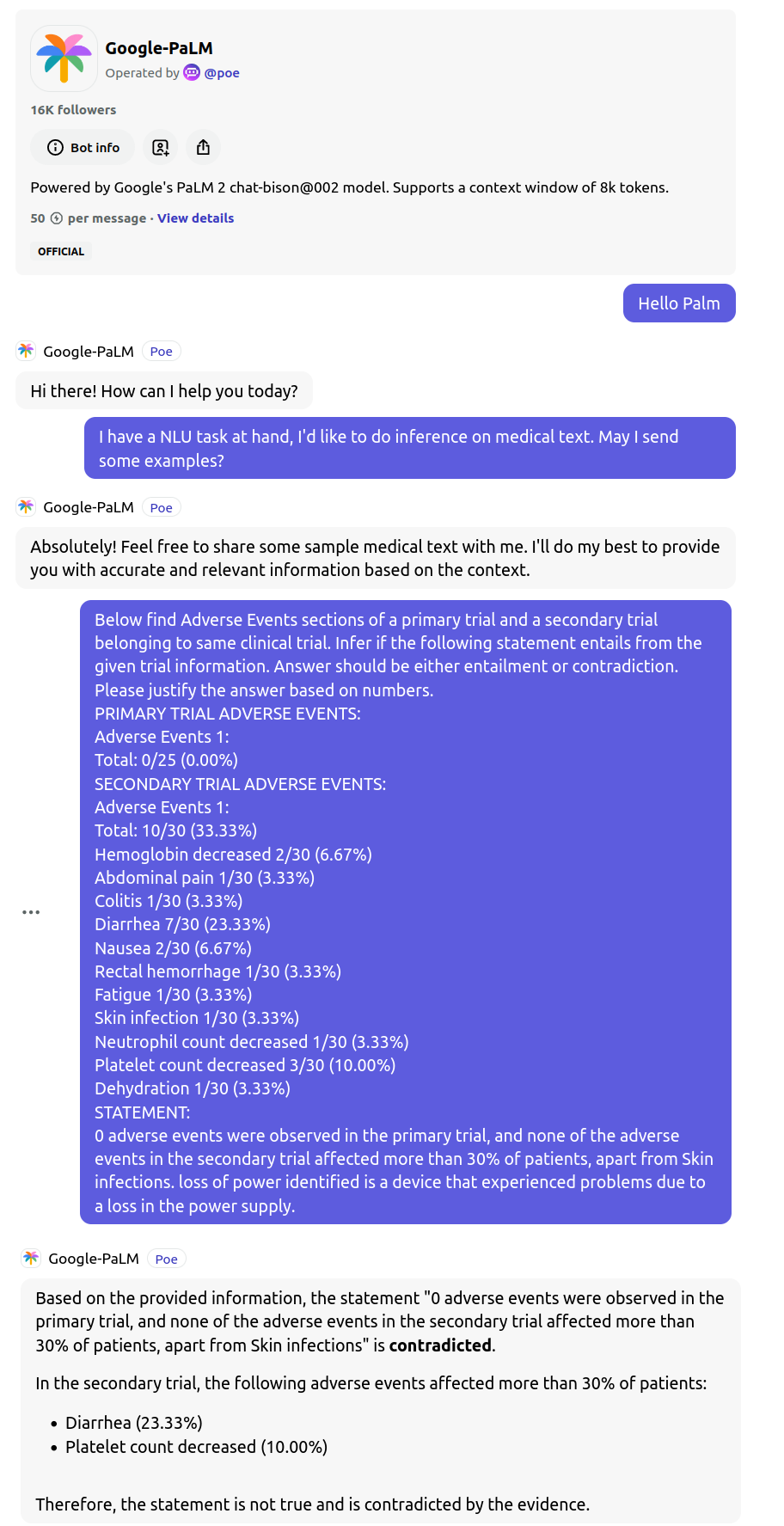}
  \caption{Initiation of the conversation with PaLM.}
  \label{fig:prompt}
\end{figure}

Figure \ref{fig:prompt} illustrates the initiation of a chat with PaLM and the method by which instances are provided during the conversation. As evident in the interaction, we maintained minimal prompting and limited additional interactions.

\section{Example Instances}
\label{sec:examples}
\begin{figure*}[t]
  \centering
  \includegraphics[width=\textwidth]{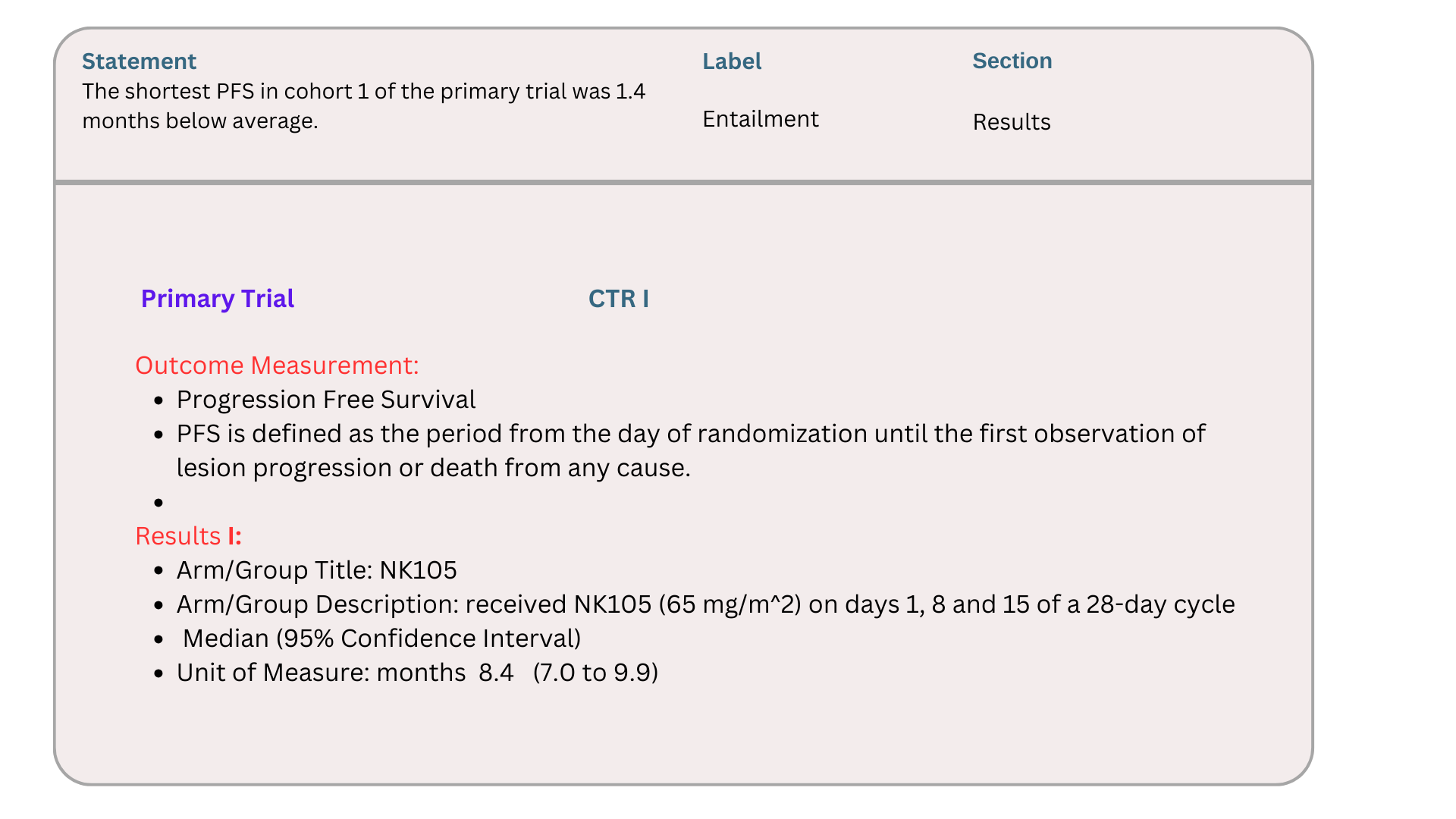}
  \caption{A challenging instance that was incorrectly predicted by the top-performing LLMs.}
  \label{fig:difficult1}
\end{figure*}

\begin{figure*}[t]
  \centering
  \includegraphics[width=\textwidth]{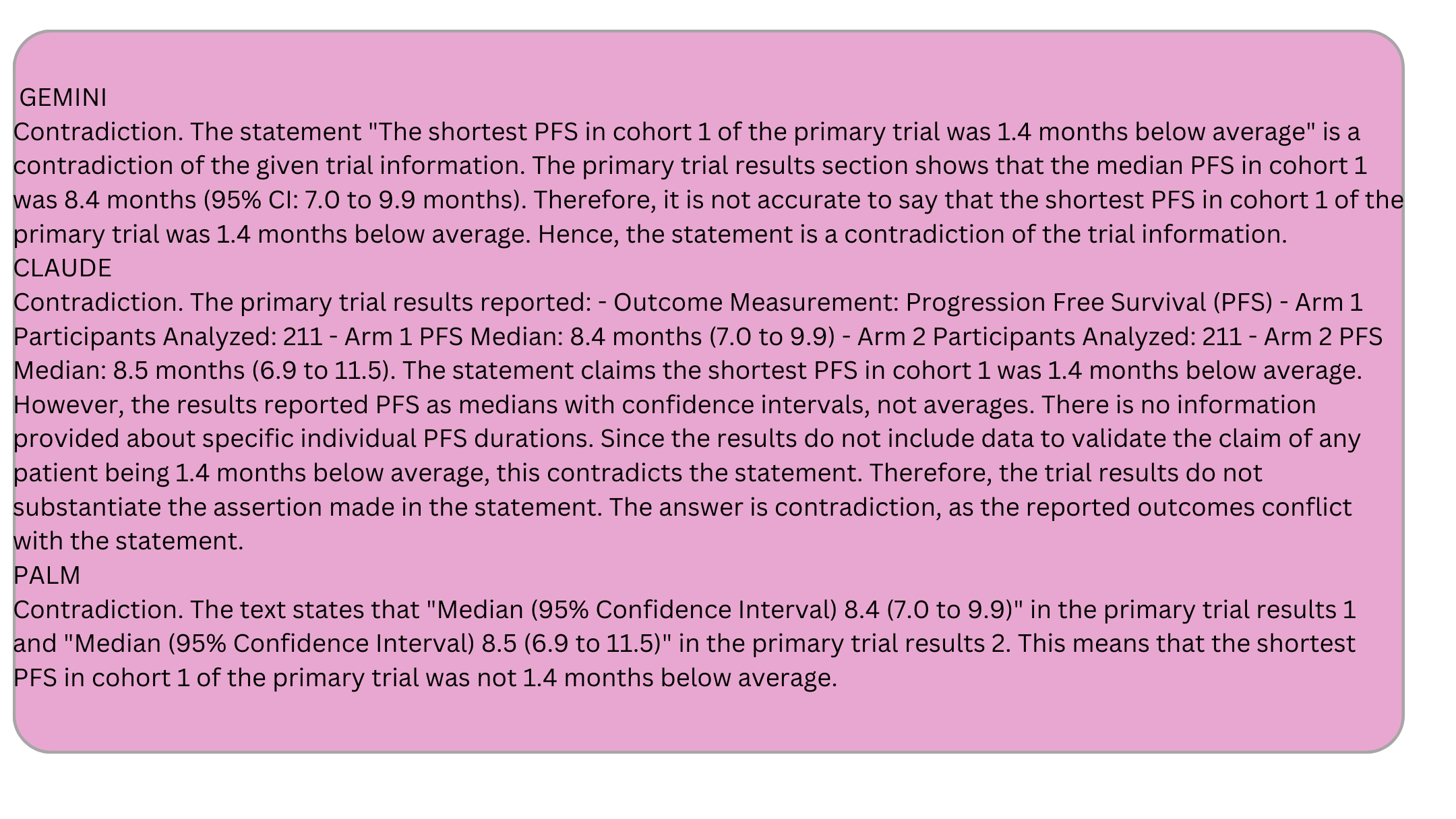}
  \caption{Responses of the top-performing LLMs to the selected challenging instance, where all models failed to exhibit any signs of numerical inference.}
  \label{fig:diff-top-answers}
\end{figure*}

\begin{figure*}[t]
  \centering
  \includegraphics[width=\textwidth]{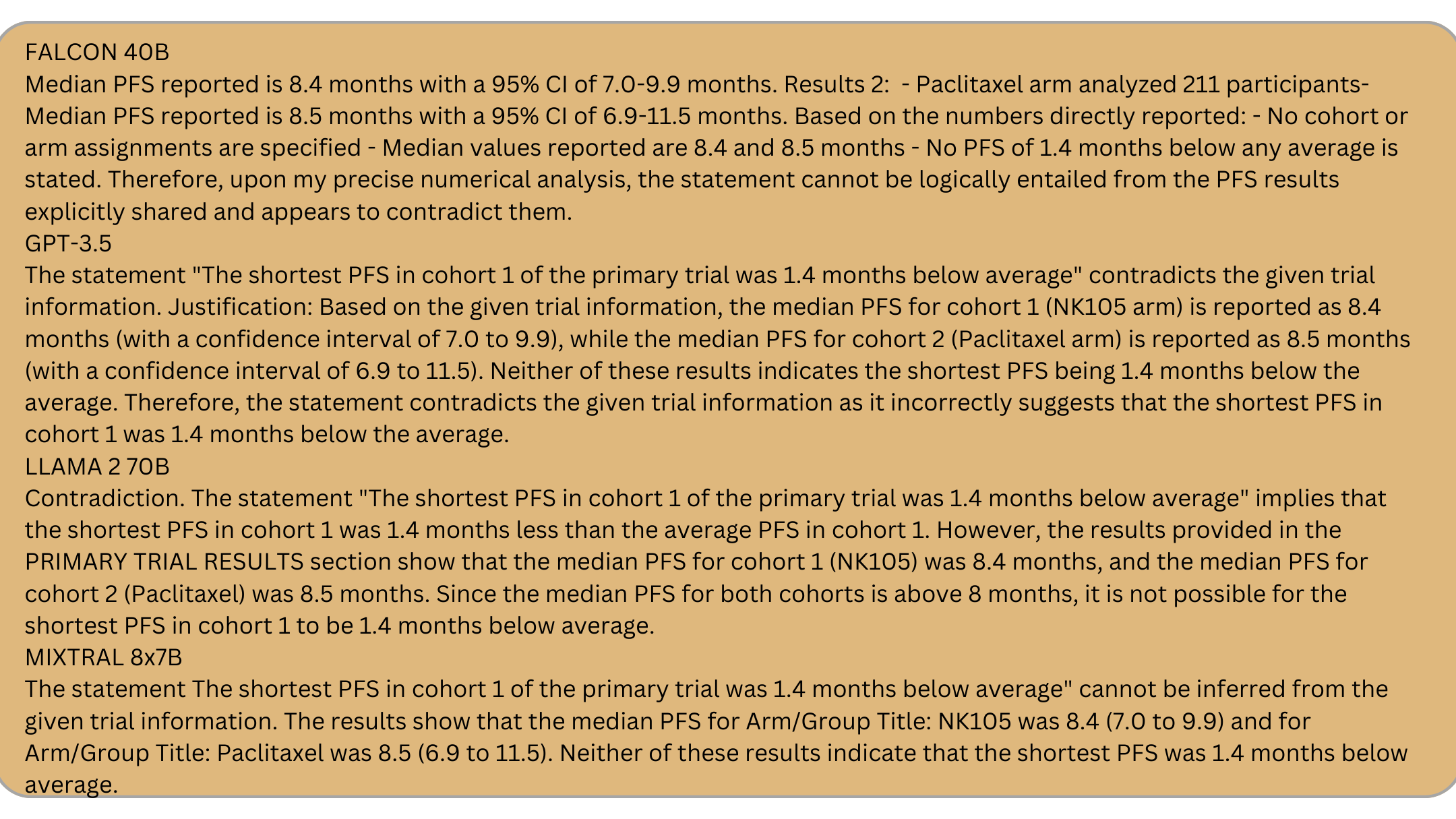}
  \caption{Responses of the low-performing LLMs to the selected challenging instance, which were not significantly different from the answers provided by the top LLMs.}
  \label{fig:diff-easy-answers}
\end{figure*}

In this section of the appendix, we present specific instances from the development set to provide readers with a concrete understanding of the performance of LLMs. Firstly, we present a challenging instance, which none of the LLMs in our study were able to correctly predict. Figure \ref{fig:difficult1} depicts this instance, which involves making an inference about the results section of a single CTR. The inference relates to PFS, a time range spanning from 7.0 to 9.9 months, with an average of 8.4 months. Consequently, the statement presents an entailment. Surprisingly, all the LLMs failed to address this instance. As depicted in Figure \ref{fig:diff-top-answers} and \ref{fig:diff-easy-answers}, the LLMs struggled to calculate the difference due to various reasons, such as difficulties in numerical deduction or becoming overly focused on linguistic details.

It is worth noting that this instance also includes an abbreviation, PFS, which is fully explained in the body of the CTR. Despite the LLMs demonstrating some understanding of this abbreviation, they ultimately failed due to their inability to perform the necessary numerical inference.

\begin{figure*}[t]
  \centering
  \includegraphics[width=\textwidth]{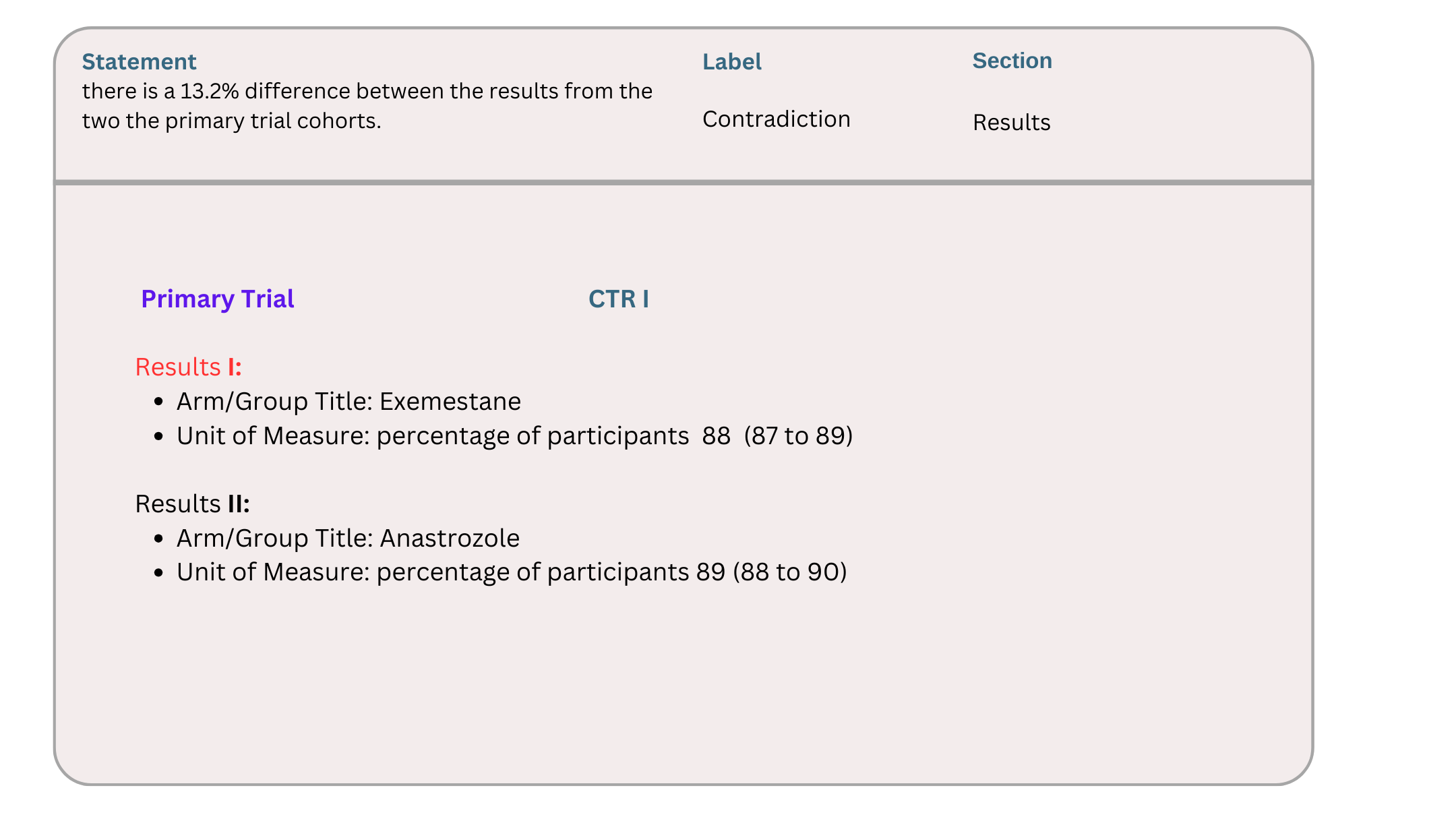}
  \caption{A numerical example extracted from the development set, focusing on inferring information from a single CTR's results section.}
  \label{fig:numeric-ctr}
\end{figure*}

\begin{figure*}[t]
  \centering
  \includegraphics[width=\textwidth]{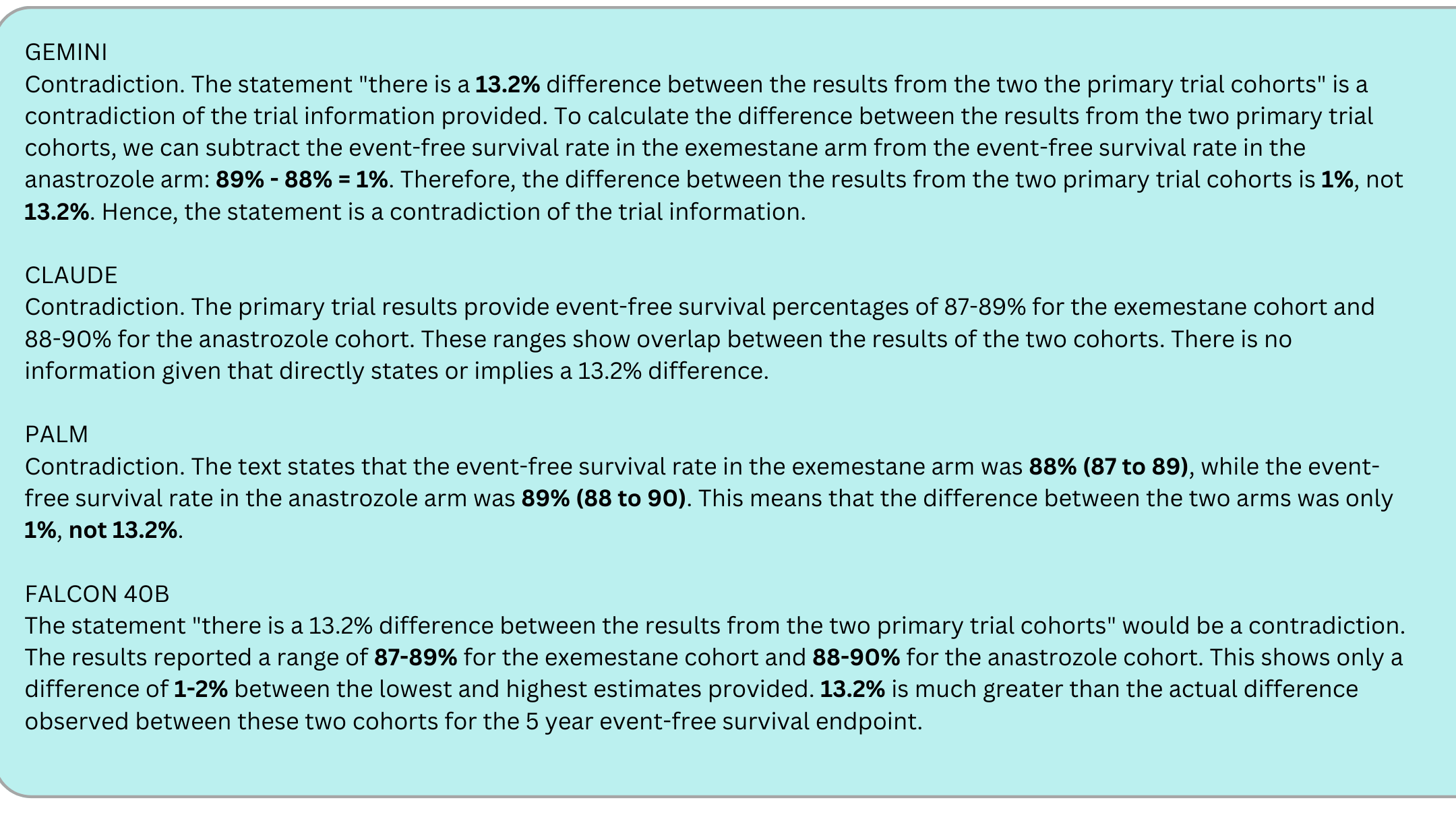}
  \caption{Responses of the top-performing LLMs to the selected numerical instance, showcasing impressive performance in numerical inference across all models.}
  \label{fig:numeric-top-answers}
\end{figure*}

\begin{figure*}[t]
  \centering
  \includegraphics[width=\textwidth]{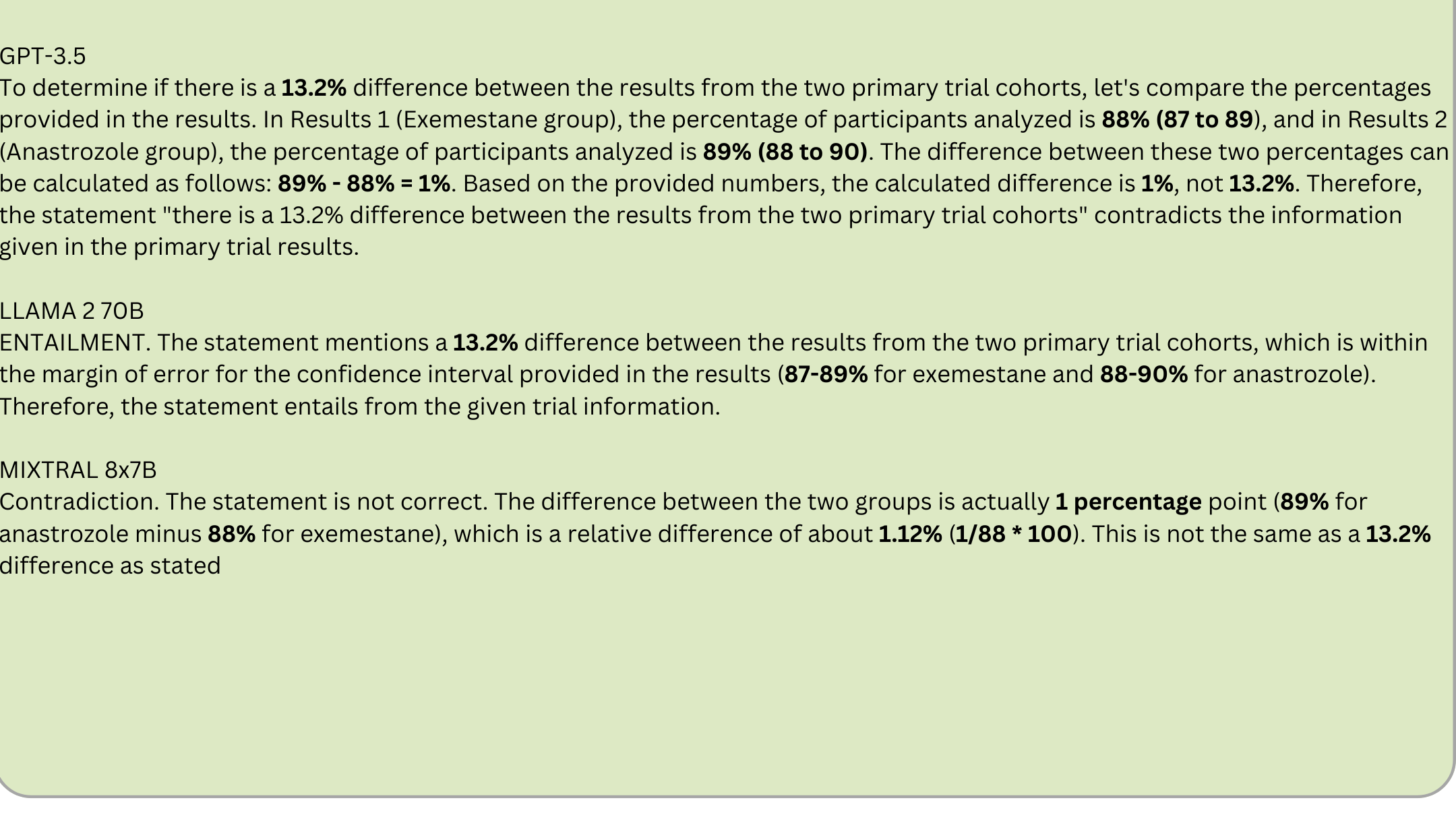}
  \caption{Responses of the bottom-performing LLMs to the selected numerical instance, where all models, except for Llama 2 70B, successfully performed the subtraction operation and made the corresponding numerical inference.}
  \label{fig:numeric-bottom-answers}
\end{figure*}

Subsequently, we present a numerical case study depicted in Figure \ref{fig:numeric-ctr} to showcase the numerical reasoning capabilities of the Language and Logic Models (LLMs). Impressively, almost all LLMs accurately predicted this particular case. However, Llama 2 70B exhibited a complete failure, displaying no signs of any numerical inference whatsoever. Figures \ref{fig:numeric-top-answers} and \ref{fig:numeric-bottom-answers} illustrate how other LLMs meticulously explained their reasoning step by step. They initiated the process by performing the subtraction $89\% - 88\% = 1\%$ and subsequently compared the result to the claimed amount of $13.2\%$.

To process numerical instances, we adopted the following approach: firstly, we utilized spaCy's Matcher component to extract all numerical expressions \footnote{\url{https://spacy.io/api/matcher}}. This component, being part of the pretrained spaCy pipelines, is incredibly helpful in extracting expressions based on patterns. These patterns can involve characteristics such as token shape, POS tags, and even entity types if the token forms part of an entity. By leveraging spaCy's built-in NER component, we could extract various numerical entity types, including cardinal numbers, ordinal numbers, percentages, and quantities. We formulated two general Matcher patterns, namely \emph{NUMERIC OP NUMERIC} and \emph{NUMERIC OP NUMERIC = NUMERIC}, and then generated all possible combinations of numerical entities and mathematical expressions by taking the cross product between numeric entity types and mathematical operator tokens. This comprehensive approach facilitated the extraction of all numerical expressions from the LLM answers. For identifying medical entities, we utilized the ScispaCy package, as medical entities are not included in spaCy's general-purpose NER models.

\begin{figure*}[t]
  \centering
  \includegraphics[width=\linewidth]{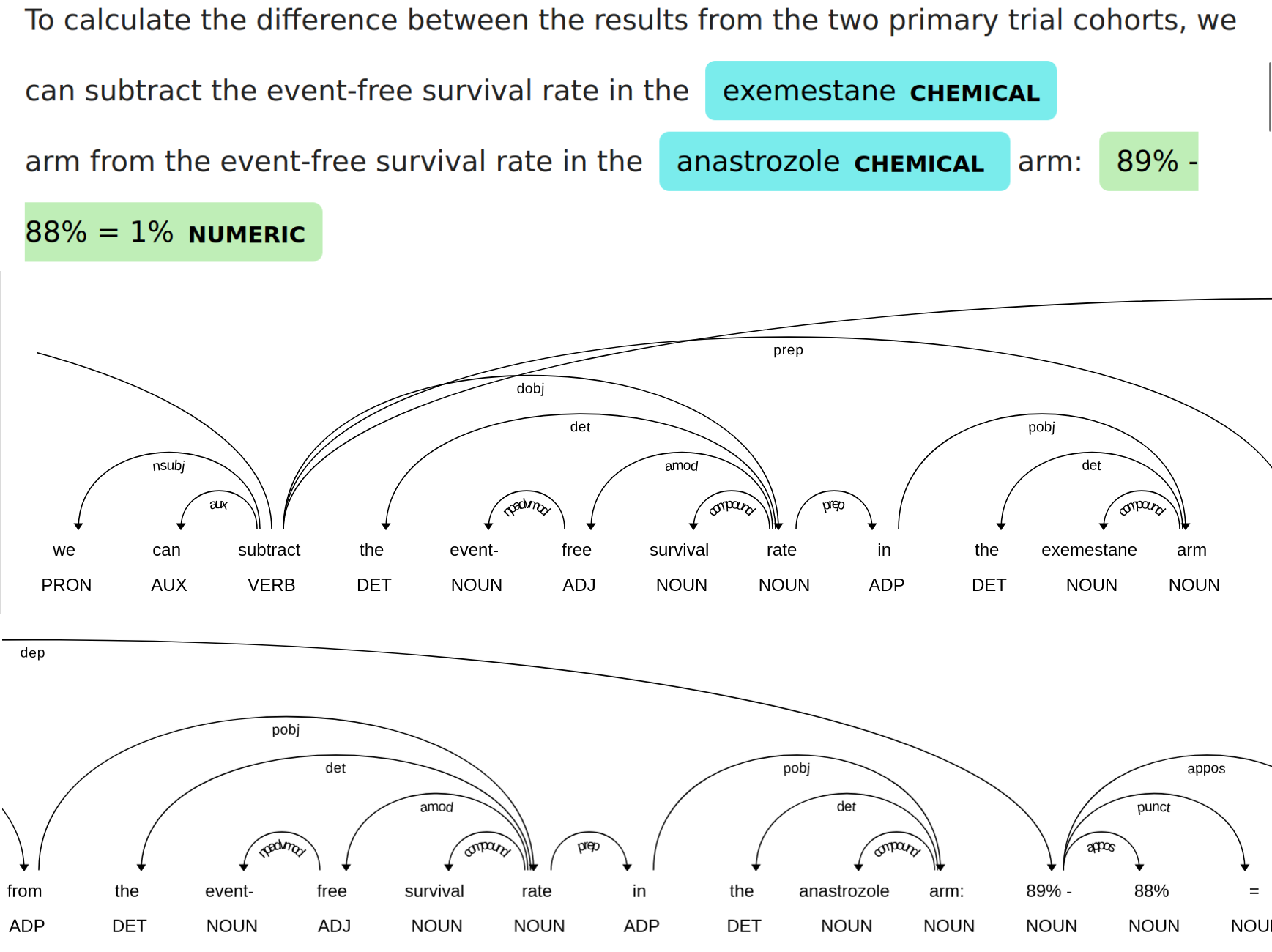}
  \caption{Semantic parse of a successful answer by Gemini. Named entities are highlighted in the above picture, where dependency tree of the sentence is exhibited in the below pictures. In the dependency tree, head token of the numerical expression \textbf{89\% - 88\% = 1\%} is \textbf{89\%} and syntactic head of \textbf{89\%} is \textbf{subtract}, which is the mathematical operation. By following the syntactic parent of the numerical expression, we reach the explanation of the chain of mathematical operations, hence we can deduce that Gemini put down a valid argument and numerical reasoning.}
  \label{fig:semantic-parse}
\end{figure*}

Following this, we parsed the dependency tree of the answer and determined the syntactic head of the numerical expression. We then examined whether the numerical expression attached meaningfully to the rest of the answer. For a detailed explanation of the reasoning process, refer to Figure \ref{fig:semantic-parse}.

\begin{figure*}[t]
  \centering
  \includegraphics[width=\textwidth]{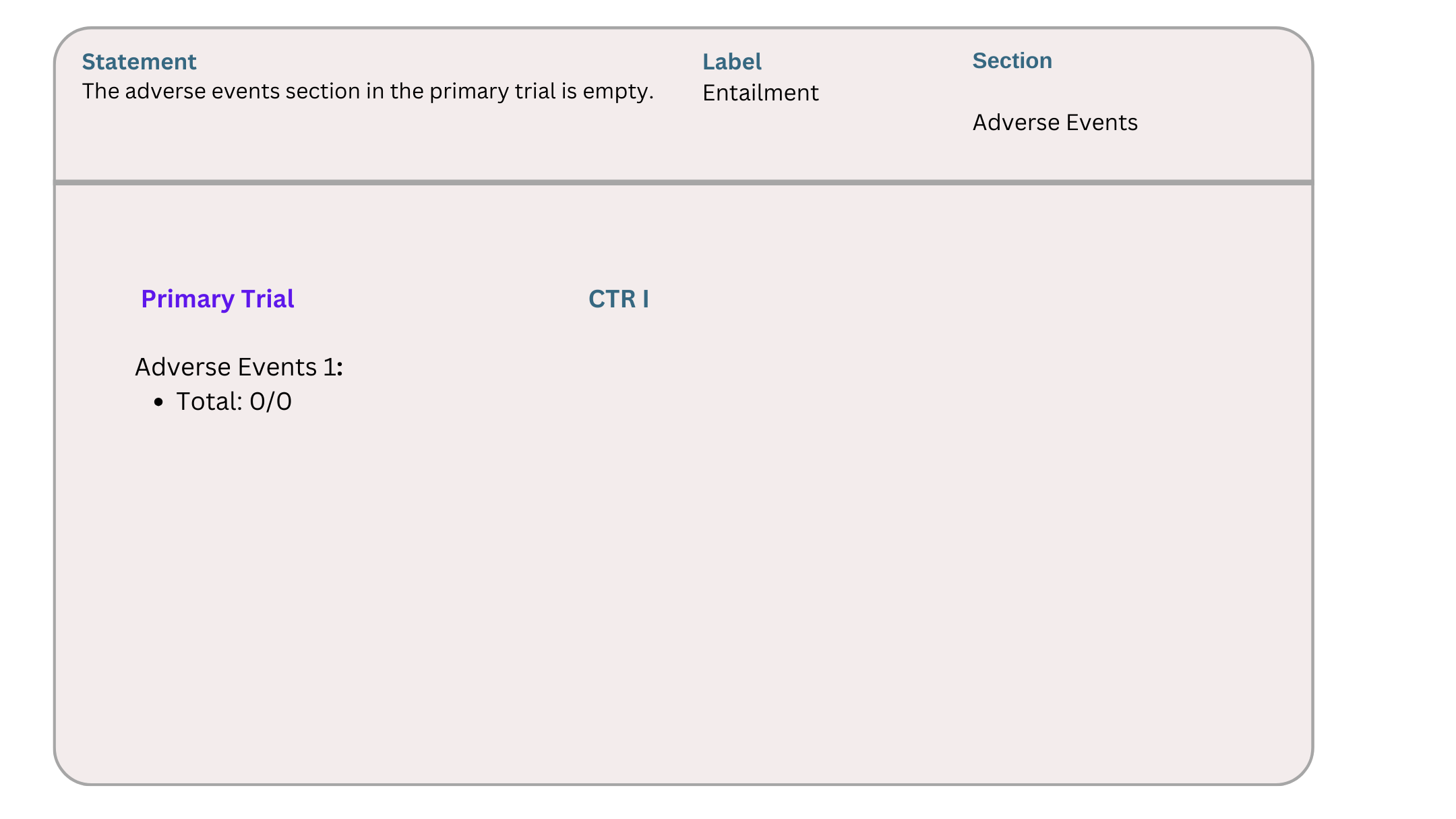}
  \caption{A CTR from the training set that is both interesting and uninteresting. It is interesting because the value of 0 denotes emptiness, but uninteresting because humans can understand 0 as representing null.}
  \label{fig:interesting-pic}
\end{figure*}

\begin{figure*}[t]
  \centering
  \includegraphics[width=\textwidth]{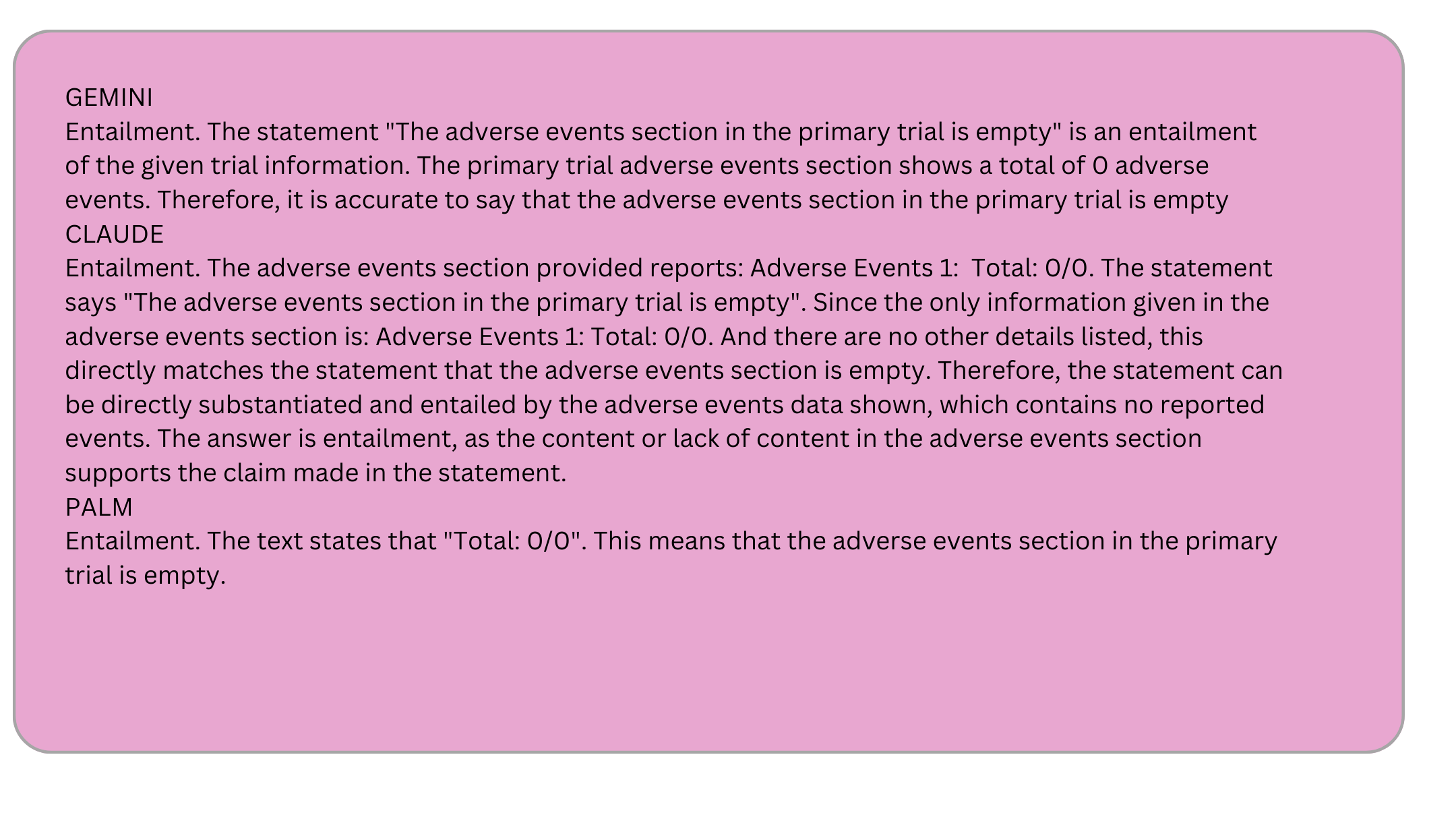}
  \caption{Responses from the top-performing LLMs, demonstrating high intelligence and deliberate reasoning.}
  \label{fig:interesting-answers}
\end{figure*}

\begin{figure*}[t]
  \centering
  \includegraphics[width=\textwidth]{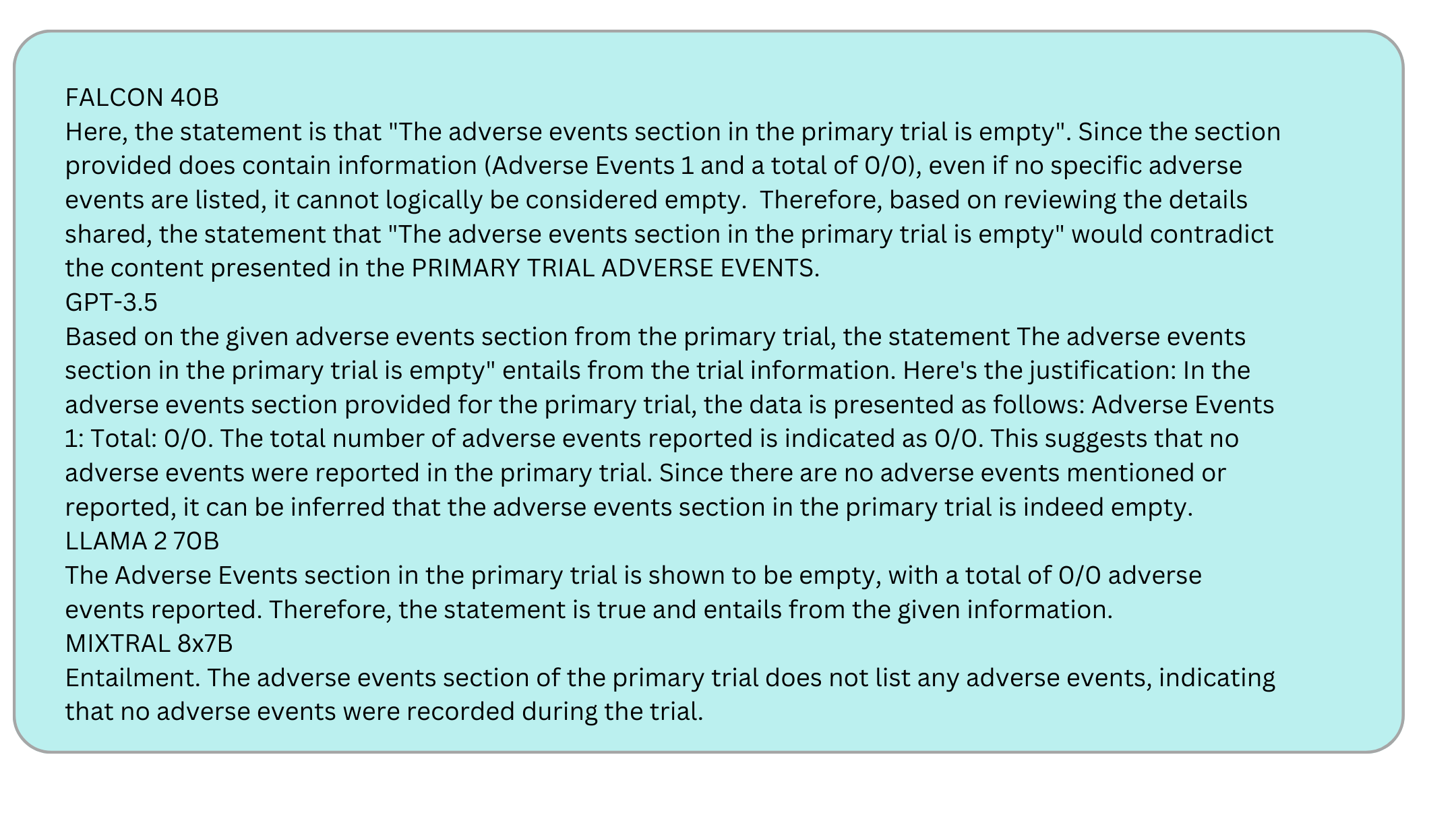}
  \caption{Responses from the low-performing LLMs, where all models, except for Falcon 40B, achieved success comparable to the top performers.}
  \label{fig:interesting-answers2}
\end{figure*}

Moving on to our list of examples, we encounter an intriguing case worth mentioning. Figure \ref{fig:interesting-pic} presents a CTR with an empty adverse events section, making it a particularly interesting example. As depicted in Figures \ref{fig:interesting-answers} and \ref{fig:interesting-answers2}, all LLMs, except for Falcon 40B, demonstrate impressive intelligence by correctly interpreting "0/0" as indicating the absence of any adverse events, thereby resulting in an empty adverse events section. This example highlights the remarkable general language understanding and common sense reasoning abilities of LLMs, transcending the boundaries of the medical domain.

\end{document}